\documentclass[10pt]{llncs}
\usepackage[T1]{fontenc}
\usepackage{graphicx}
\usepackage{hyperref}
\hypersetup{colorlinks,allcolors=black}
\usepackage{url}
\usepackage{times}
\usepackage{amsmath}
\usepackage{amssymb}
\usepackage{bbm}
\usepackage{subcaption}
\usepackage[rightcaption]{sidecap}
\usepackage{float}
\usepackage{longtable}
\usepackage{multirow}
\usepackage{xcolor}
\usepackage{svg}
\usepackage[off]{svg-extract}
\svgsetup{clean=true}

\usepackage[ruled,vlined]{algorithm2e}
\definecolor{commentcolor}{RGB}{110,154,155}
\newcommand{\PyComment}[1]{\ttfamily\textcolor{commentcolor}{\# #1}}  
\newcommand{\PyCode}[1]{\ttfamily\textcolor{black}{#1}}
\usepackage{color}

\begin{document}
%

\title{Self-Reinforcement Attention Mechanism For Tabular Learning }
\author{ Kodjo Mawuena Amekoe$^{1,3}$, Mohamed Djallel Dilmi $^{1,2}$, Hanene Azzag$^{1}$, Mustapha Lebbah$^{1,2}$,\\ Zaineb Chelly Dagdia$^{2}$ and  Gregoire Jaffre$^{3}$
%
%
%
\vspace{.3cm}\\
%
1- Sorbonne Paris Nord University - LIPN, UMR CNRS 7030, Villetaneuse, France
%
\vspace{.1cm}\\
2- Paris-Saclay University - DAVID Lab, UVSQ, Versailles, France 
\vspace{.1cm}\\
3- Groupe BPCE, 7 promenade Germaine Sablon 75013 Paris, France \\
}
\institute{}
\maketitle 
\begin{abstract}
Apart from the high accuracy of machine learning models, what interests many researchers in real-life problems (e.g., fraud detection, credit scoring) is to find hidden patterns in data; particularly when dealing with their challenging imbalanced characteristics. Interpretability is also a key requirement that needs to accompany the used machine learning model. In this concern, often, intrinsically interpretable models are preferred to complex ones, which are in most cases black-box models. Also, linear models are used in some high-risk fields to handle tabular data, even if performance must be sacrificed.
In this paper, 
we introduce Self-Reinforcement Attention (SRA), a novel attention mechanism that provides a relevance of features as a weight vector which is used to learn an intelligible representation. This weight is then used to reinforce or reduce some components of the raw input through element-wise vector multiplication.  
Our results on synthetic 
and real-world imbalanced
data show that our proposed SRA block is effective in end-to-end combination with baseline models.
\keywords{Attention  \and Interpretability \and classification.}
\end{abstract}
\vspace{-0.5cm}
\section{Introduction}
While deep learning models continue to provide impressive performance for computer vision and Natural Language Processing (NLP), their practical use in some real-life  problems (e.g. credit scoring) remains limited due to legislation with one of the important points: the interpretability of the used models  (e.g., GDPR: Article 22 in Europe).
Since the promising results of the transformer architecture on machine translation tasks \cite{VanillaTransformer}, many efforts have been made to improve models' accuracy for tabular modeling using the attention mechanism \cite{SAINT,NPT,TabTransformer,FTransformer}. The principal motivation behind our work is to push forward this effort by proposing an interpretable attention model for tabular learning. 
To achieve this goal, we found it necessary to develop a representation learning block or layer that (i) preserves the initial feature space (i.e., $\mathbbm{R}^{p} \longrightarrow \mathbbm{R}^{p}$), and (ii) reduces to the maximum some extra steps (e.g., residual connection, LayerNorm) that make the overall architecture less interpretable.
\\In this paper, we present a new attention-based representation learning block for tabular data, called Self-Reinforcement Attention (SRA). 
We summarize our contributions as follows:
\begin{itemize}
    \item SRA is a pre-processing block that allows to learn intelligible representation by weighting each raw feature with a positive alignment (i.e., an attention score). The obtained representation, called ``reinforced vector'', is then passed to an aggregation model to provide the decision boundary. 
    It is based on the previous work about Epigenetics algorithms \cite{dilmi2023epigenetics}.
    \item SRA provides a feature score that facilitates interpretations. This score is then used as a coefficient to amplify or reduce some characteristics according to the context  of the observations (here the spatial information), allowing to: (i) take into account possible interactions without the need to artificially add additional features or terms, (ii) identify important features for the final prediction.  
 \item Our experiments on synthetic and benchmark imbalanced datasets show a promising  performance while preserving  intrinsic interpretability of the resulting architecture. 
\end{itemize}
\indent The rest of the paper is organized as follows: Section \ref{RelatedWork} presents a brief discussion of state-of-the-art works. Section \ref{SRAsection} describes the SRA block and its architecture. The experimental setup, the discussion of the obtained results and the limitations are presented in Section \ref{ExperimentsSection}. Finally, Section \ref{ConclusionSection} concludes the paper and highlights some perspectives.
%
\section{Related work}
\label{RelatedWork}

\textbf{Classical models.}
For many tabular or heterogeneous data classification problems, tree-based ensemble models are still widely used and preferred as they generate high performance results while not being greedy in terms of computational resources. Random Forest  \cite{RandomForest} is one of the well-known models that combines the prediction of several independent trees through majority voting or average. Tree based models are also used in a sequential manner (Gradient Boosting Machine or GBM) where the newest trees are built to minimize the errors of the previous ones. XGBoost  \cite{Xgboost} and LightGBM \cite{LightGBM} are two  fast implementations of GBM that empower the regularization in the boosting mechanism and are considered as the state-of-the-art for many problems or competitions. As for linear models, these are commonly used in settings where models’ interpretability is strongly required. For these models, linear relations between features are expected otherwise continuous variables are discretized aiming to take into account non-linearity; otherwise other features are added as interaction terms. \\
\\
\textbf{Deep learning models and  feature attribution.}
Many deep learning models were designed to provide local explanations to their predictions. Among these models, we mention  Neural Additive Models (NAM) \cite{NAM} which present a neural implementation of the classic Generalized additive models (GAM). In addition to the marginal feature contribution, NAM provides the shape function which by visualization can help understand the global contribution of a given feature. NODE-G$^2$AM   \cite{NODE_GAM} is an improvement of NAM built on top of a NODE architecture \cite{NODE} to take into account pairwise interactions among features. Among the drawbacks of these deep learning architectures is that they apply one sub-network per feature (or pair of features) which can be very resource consuming, especially for high dimensional problems, and   the management of higher order interactions is not guaranteed.
In TabNet \cite{Tabnet}, a multiplicative sparse mask vector is used for instance-wise feature selection and the masks are aggregated across many stages to compute features’ importance. Matrix multiplication trick is used to reconstruct initial feature dimension which is lost with the passage in the feature transformer module. Contrary to TabNet, our proposed SRA model preserves the initial feature dimension and uses the dot-product to compute its attention weights.\\
\\
\textbf{Attention-based models for tabular data classification.}
Compared to classical models, deep learning models have several advantages in various settings, mainly (i) in continuous learning especially in the presence of concept drift (e.g., using transfer learning or fine tuning), (ii) in multimodal problems (e.g., encode tabular information with text, image, etc.), and (iii) in multitask settings \cite{NAM}. These reasons motivated many researchers to improve the traditional fully connected MultiLayer Perceptron (MLP) by imitating tree models \cite{NODE} or to use the attention mechanism  \cite{SAINT,NPT,TabTransformer,FTransformer}. One common feature of  these attention-based architectures is that each feature is considered  as a token and is embedded similarly as in \cite{VanillaTransformer}. 
This consideration, although favoring the expressiveness of the resulting architecture, leads to a significant increase in the number of learnable parameters and complicates the explanation of local predictions, especially when the number of transformer layers or stages exceeds 1. The SRA-based models proposed in our work use less parameters compared to \cite{SAINT,NPT,TabTransformer,FTransformer}. Also they are accurate in comparison to state-of-art models while being self-explainable, in the sense that they provide an intrinsic explanation of their predictions.
%
\section{Self-Reinforcement Attention (SRA)}
\label{SRAsection}


%
\subsection{SRA Model }
The challenge in most supervised tabular learning problems using attention mechanism \cite{SAINT,NPT,TabTransformer,FTransformer} is to estimate the output $\hat{y} = f_{\theta}(\mathbf x)$ given the feature vector $\mathbf x = (x_1,...,x_p) \in \mathbbm{R}^p$. The parametric model $f_{\theta}$ is learned using the training data $\mathcal{D}=\{(\mathbf x_i, y_i )\}_{i=1}^{n}$ with  $y_i \in \{0,1\}$ for binary classification or $y_i \in \mathbbm{R}$ for regression tasks. 
Our proposed SRA model $f_{\theta}$ (Fig~\ref{fig:sramodel}) contains a SRA block  (Fig~\ref{fig:srablock}) which is a novel attention mechanism layer denoted as a function $a(.)$. Given the raw input $\mathbf x$, the SRA block produces an attention vector 
$\mathbf a = (a_1,...,a_i,...a_p)$. Thereafter the attention vector is used to learn an intelligible representation $\mathbf o =(o_1,...,o_i,...,o_p)$ 
as follows:
\begin{equation}\label{eqn:reinforcedrepresenattion}
\mathbf o = \mathbf a\odot \mathbf x 
\end{equation} 
where $\odot$ is the element-wise multiplication. \\
If we instantiate our supervised aggregation model (Figure \ref{fig:sramodel}) as a linear transformation, then the SRA model can be formalized as follows:
\begin{figure}[h]

 \begin{subfigure}{0.52\textwidth}
 \includegraphics[width = 0.99\textwidth,height=5cm]{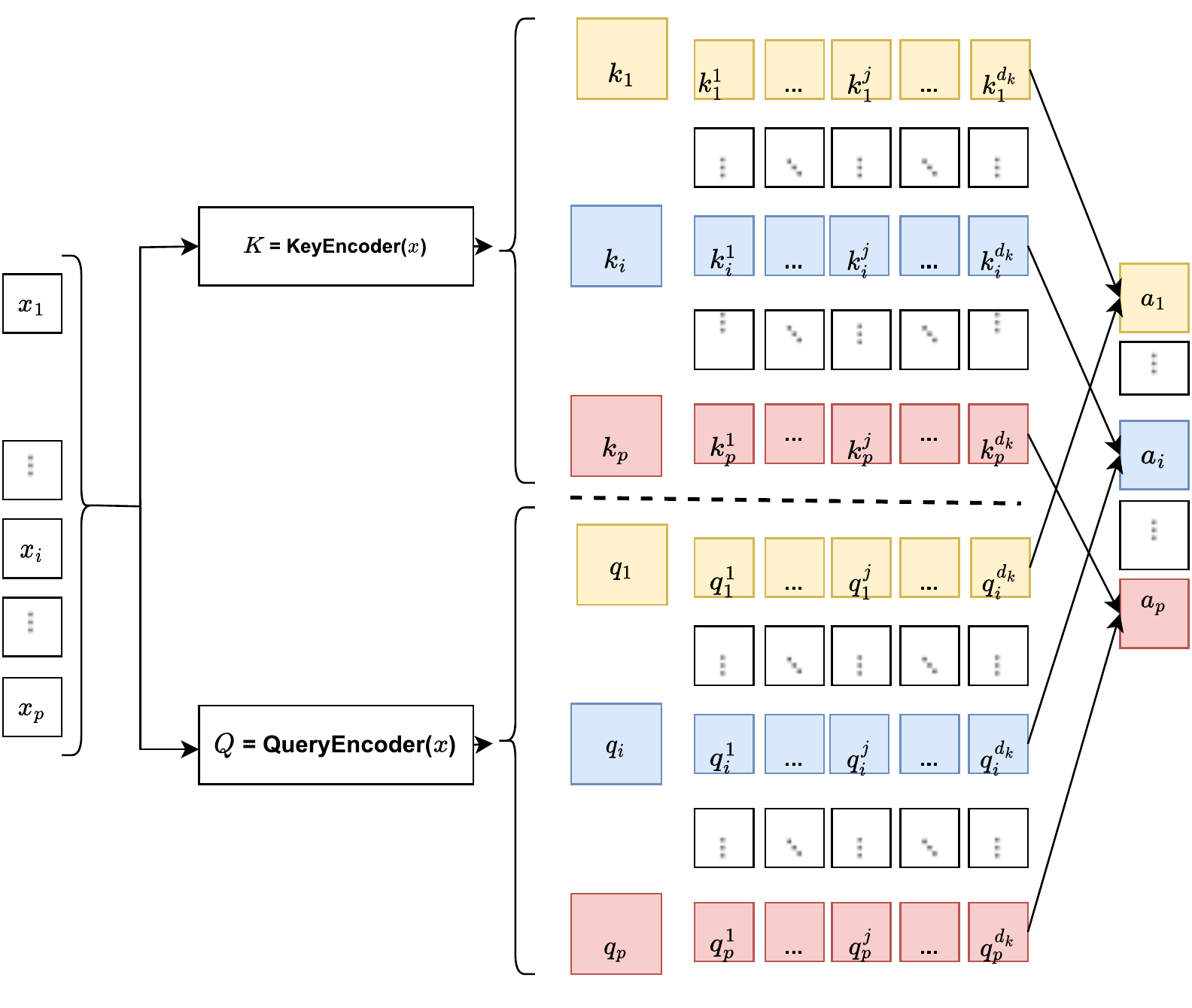}
        \caption{SRA Block. The $KeyEncoder$ (resp. $QueryEncoder$) produces directly $p$ keys (resp. queries)}
        \label{fig:srablock}
    \end{subfigure}
          \hspace{0.5cm}
  \begin{subfigure}{0.48\textwidth}
 \includegraphics[width = 0.99\textwidth,height=5cm]{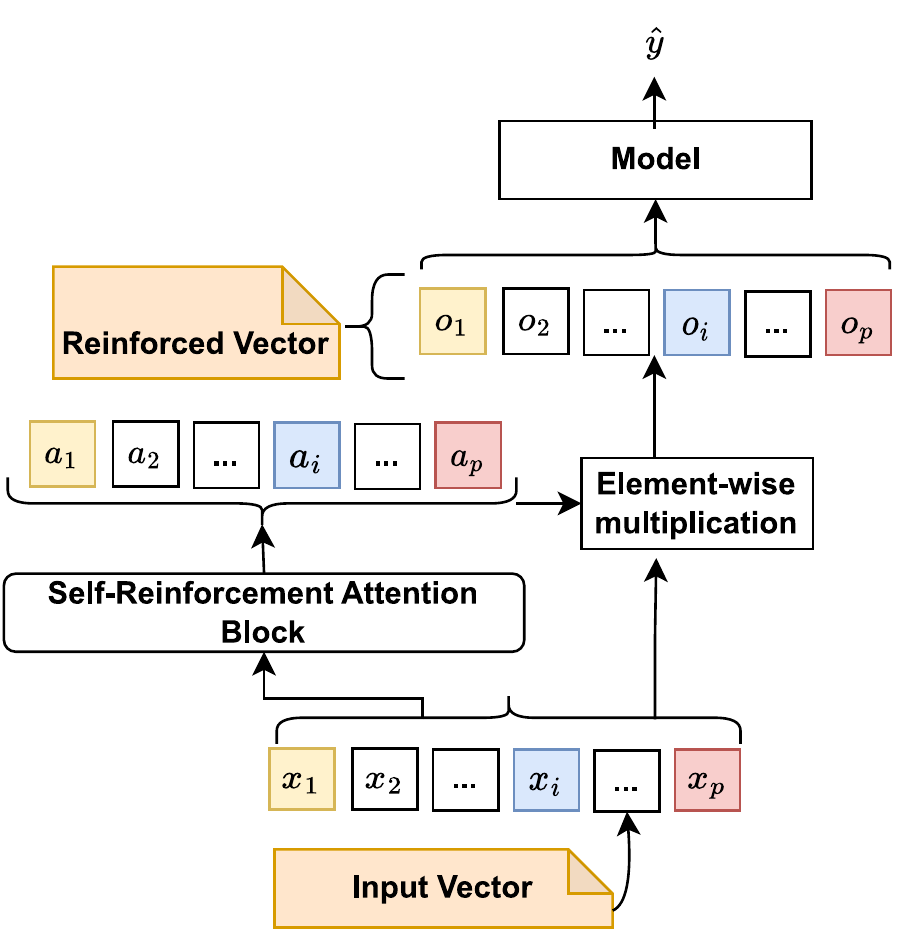}
        \caption{SRA model. The attention vector $\mathbf a =(a_1,...,a_p) \in \mathbbm{R}^p$ provided by the SRA block is used to produce a \textit{reinforced} vector $\mathbf o =(o_1,...,o_p) \in \mathbbm{R}^p$}
        \label{fig:sramodel}
    \end{subfigure}
    \caption{ SRA architecture. }
    \label{SRAarchi}
\end{figure}
%
\begin{equation} 
\label{eqn:SRA}
\begin{split}
g(\hat{y}) &= \boldsymbol\beta \cdot \mathbf o \\
   &= \beta_1 o_1 +...+\beta_i o_i+...+\beta_p o_p \\
   &= \beta_1 a_1 x_1 +...+\beta_ia_ix_i+...+\beta_pa_px_p \\
\end{split}
 \end{equation}
$\beta_ia_ix_i$ represents the contribution (the prediction importance) of the feature $x_i$ to the output, $\boldsymbol\beta = (\beta_1, \beta_2,...,\beta_p)$ is the linear regression coefficients and
 $a_i$ is interpreted as the amplification (or the correction) that the feature $x_i$ received from other features or itself due to the interactions. We call this SRA model (instantiation) as SRALinear.  $g$ represents the link function (e.g., usually $g(\mu)= \log(\frac{\mu}{1-\mu})$ for binary classification and $g = Identity$ for regression tasks). 
\vspace{-0.2cm}
\subsection{SRA block}
\vspace{-0.05cm}
Given the input vector $\mathbf x = (x_1,...x_i,...,x_p)\in \mathbbm{R}^p$, the SRA block encodes it  into $p$ keys in $K = [{\mathbf k}_1,{\mathbf k}_2,...,{\mathbf k}_i,...,{\mathbf k}_p]^T$ with ${\mathbf k}_i=(k_i^1,..., k_i^{d_k}) \in \mathbbm{R}^{d_k}$ using the key encoder and queries matrix $Q = [{\mathbf q}_1,{\mathbf q}_2,...,{\mathbf q}_i,...,{\mathbf q}_p]^T$ with ${\mathbf q}_i=(q_i^1,..., q_i^{d_k}) \in \mathbbm{R}^{d_k}$ using the query encoder (see Figure \ref{fig:srablock} and the 
pseudocode provided in Algorithm \ref{PytorchSRAPseudo}).
%
%
The matrix of queries ($Q$) and keys ($K$) are generated by two separate fully connected feed-forward networks ($FFN$)
namely $QueryEncoder$ and $KeyEncoder$. \\
\indent The $KeyEncoder$ (resp. $QueryEncoder$) produces directly $p$ keys (resp. queries) using a single $FFN$ instead of using $p$ independent $FFN$s per feature as in \cite{SAINT,TabTransformer}.
This embedding should be particularly useful for heterogeneous (tabular) data especially in the presence of strong features' interactions and at the same time alleviate the need of using several attention blocks (layers) or extra processing which could affect the interpretable  of the attention coefficients. Furthermore, with a Sigmoid activation function, all elements $k_i^j$  of $K$ (resp. $q_i^j$ of $Q$) are scalar numbers bounded in $[0,1]$.\\
\indent The keys in $K$ are compared to the queries $Q$ component by component, allowing to quantify the alignment of different transformations of the same input calculating the attention weights $\mathbf a=(a_1,..,a_i,...,a_p)$
as follows :
\begin{equation}
      {a}_i = \frac{{\mathbf q}_i\cdot {\mathbf k}_i}{d_k} \quad \text{for} \quad i \in {1,\cdots, p}
\label{eqn:dotproduct}
 \end{equation}
 We further use the scaling by ${d_k}$ in order to reduce the magnitude of the dot-product and to get dimension-free attention coefficients $a_i \in [0, 1]$.\\
 \indent We propose this attention estimation to produce a concise explanation of the decision process.
 Indeed, considering the potential internal conflict between the input components (due to the interactions), the attention weights vector $a$ may enhance or reduce some components (of the input vector) at strategic and specific positions.
 %
%
\vspace{-5mm}
 \begin{algorithm}[h]
\SetAlgoLined
    \PyComment{$b$ is batch size, $p$ the number of features } \\
    \PyCode{def forward(self, x):} \\
    \Indp   
        \PyCode{Q = self.KeyEncoder(x)} \PyComment{Q is $(b, p, d_k$)} \\ 
        \PyCode{K = self.QueryEncoder(x)} \PyComment{K is $(b, p, d_k$)} \\
        \PyCode{QK = Q*K*self.scale} \PyComment{scale$= 1/d_k$, QK is $(b, p, d_k$)} \\
        \PyCode{a = QK.sum(axis = -1)} \PyComment{a is $(b, p$)} \\
        \PyCode{return a}\\
    \Indm 
\caption{PyTorch-style forward pass pseudocode of the SRA Block}
\label{PytorchSRAPseudo}
\end{algorithm}
\vspace{-5mm}
\section{Experiments}
\label{ExperimentsSection}
\subsection{Experimental setup}
\label{Experimentalsetups}
\label{ExperiencesDetails}
Our motivation when building the SRA is to combine interpretability and performance in a single model with a focus on imbalanced classification tasks. Typically, the interpretability of models is assessed separately from their performance, which can make it challenging to gauge the effectiveness of our SRA solution. Nonetheless, we believe it is appropriate to measure the value of our SRA solution by comparing it to both comprehensible and fully complex benchmarks using the following criteria:
\begin{itemize}
\item  \textbf{Intelligibility}:  Are the representations learned by  SRA understandable?  Are the explanations provided by   SRA models faithful? 
\item  \textbf{Effectiveness}: Are SRA-based models accurate compared to state-of-the-art models? 
\end{itemize}
\paragraph{\bf Datasets.}
As we focus particularly on finance as an application domain, we considered three UCI datasets (Default of Credit Card Clients, Bank Marketing, and Adult Income) and four Kaggle datasets (Credit Card Fraud, Bank Churn Modelling, Blastchar, and Telco Churn) and the Heloc Fico dataset for our experiments. All of these datasets are used for binary classification tasks, and the number of features (both numerical and categorical) ranges from 10 to 63. The percentage of positive class instances varies between 0.17\% and 48\% (see Table \ref{StatDatasets} for further details).
%
Unless otherwise specified, all categorical inputs are one-hot encoded, and numerical inputs are scaled using the mean and standard deviation
to accelerate the convergence of the algorithms.
\begin{table}[t]
\begin{center}
\caption{Benchmark datasets } 
\label{StatDatasets}
    \begin{tabular}{c c c c c}
    \hline
     Datasets & $\#$ Datapoints & $\#$  features & $\#$  Categorical features & Positive Class (\%) \\ 
     \hline
      Bank Churn  & 10000 & 10 & 2 & 20.37\\
      \hline 
      Credit Default & 30000 & 23 & 3 & 22.16\\
     \hline 
      Bank Marketing & 45211 & 16 & 9 & 11.70 \\
      \hline 
      Adult Income & 30162 & 14 & 8 & 24.89\\
      \hline
      Credit Card Fraud & 284807 & 29 & 0 & 0.17\\ 
      \hline
     Blastchar & 7043 & 19 & 16 & 26.54\\ 
      \hline
     Telco Churn & 66469 & 63 & 0 & 20.92 \\ 
      \hline
     Heloc Fico & 10459 & 23 & 0 & 47.81\\ 
      \hline
    \end{tabular}
\end{center}
\vspace{-6mm}
\end{table}

\paragraph{\bf Model setup.}
\begin{itemize}
\item {Choice of the query and key encoder:}
we use the same architecture  for the key and query encoders which is a two hidden layers fully connected neural network of dimension $\{d_1, d_2\}$ with, $d_1 = p\times (d_k/4)$ and $d_2 = p\times (d_k/2)$, $d_k\ge 4$. 
\item {Regularization:}
to increase the generalization power, we used regularization in the SRA block. Specifically, we used dropout \cite{Dropout} in both the key and query encoders during the training. Also, we used weight decay ($L_2$ penalization) to empower the smoothness in the embeddings (of the key and query). 
\end{itemize}
\paragraph{\bf Evaluation measures.}
\label{EvalMetrics}
We evaluate the models using 5-stratified fold cross validation (80\% for the training) and report the mean and standard deviation of the Area Under the ROC curve (AUCROC) on the validation set. Particularly for highly imbalanced datasets (e.g., the Credit Card Fraud dataset), we optimize and report the Average Precision or Precision-Recall (AUCPR). In fact, AUCPR gives a more informative picture of an algorithm's performance than AUCROC in highly imbalanced data settings and algorithms that optimize AUCPR are guaranteed to optimize AUCROC \cite{aucrocVsaucpr}.\\
\vspace{-0.50cm}
\subsection{Intelligibility of SRA}
%
\label{ExplanationsSRALinear}
One interesting property of an SRA-based model is that it provides interpretable information about its behavior. In this section, we explore some of these interpretable aspects through visualizations and the ability to identify relevant features. We focus in this section  on its combination with the linear model.
The SRALinear model (Equation \ref{eqn:SRA}) has two interesting properties:
\begin{enumerate}
    \item Each feature $x_i$ appears in the equation as in a classical linear model and $\beta_ia_ix_i$ is its contribution to the output.
    \item Faithfulness: the attention coefficients are clearly correlated to model's outputs. This is actually a desirable property for considering attention as an explanation of predictions \cite{AttentioNotExplantion}.
\end{enumerate}
%
%
%

\begin{figure}[t]
    \begin{subfigure}{0.50\textwidth}
        \includegraphics[width = 0.90\textwidth,height=3cm]{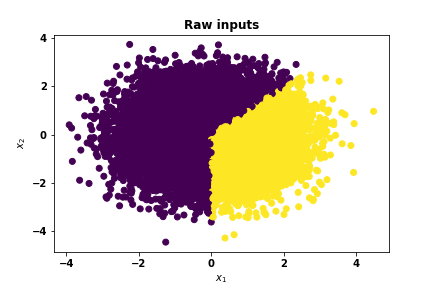}
        \caption{{Original input data}}
        \label{fig:SRADemooriginal}
    \end{subfigure}
    \begin{subfigure}{0.5\textwidth}
        \includegraphics[width = 0.90\textwidth,height=3cm]{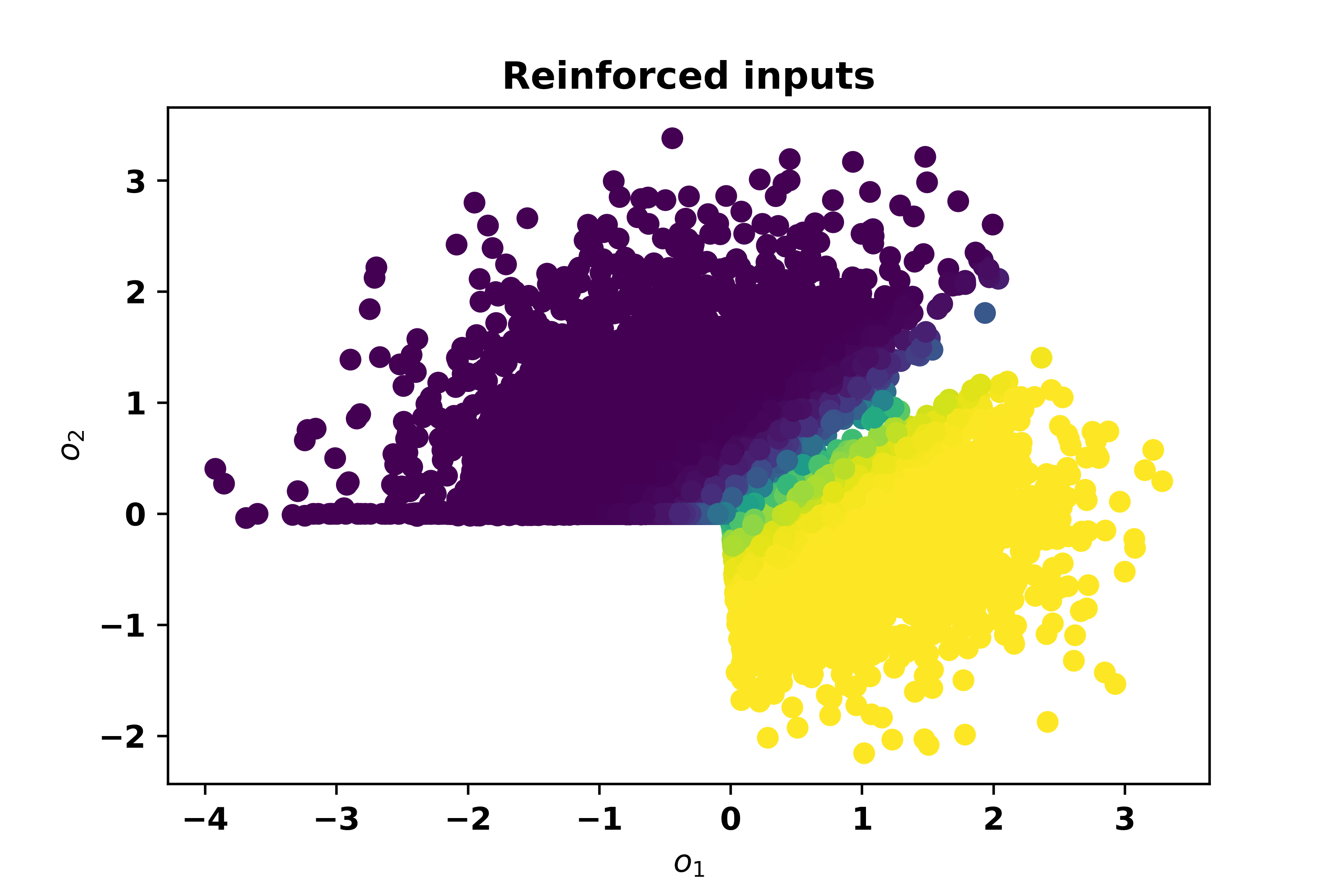}
        \caption{{Learned reinforced inputs with SRA}}
        \label{fig:SRADemoSra}
    \end{subfigure}
    \caption{Illustration of the reinforcement process on 7500 synthetic data points  with 0 mean, unity variance Gaussian distribution. 
    The yellow color is used for the class of interest. 
    The green color a possible decision boundary to separate the two classes.}
    \label{SRADemo}
\end{figure}

\begin{figure}[t]
    \begin{subfigure}{0.50\textwidth}
        \includegraphics[width = 0.90\textwidth,height=3cm]{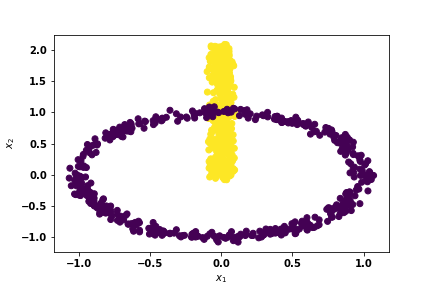}
        \caption{{Original input data}}
                \label{SRADemoChainLinkOriginal}
    \end{subfigure}
    \begin{subfigure}{0.5\textwidth}
        \includegraphics[width = 0.90\textwidth,height=3cm]{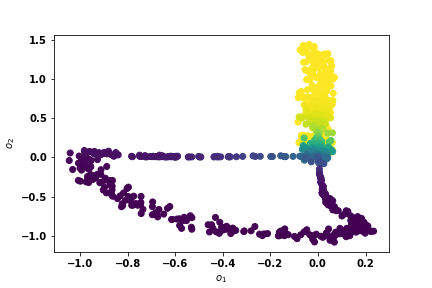}
        \caption{{Learned reinforced inputs with SRA}}
        \label{SRADemoChainLinkSRA}
    \end{subfigure}
    \caption{Illustration of the reinforcement process on the chainLink 2D\cite{Clustering_2D_datasets}  with 1000 datapoints}
    \label{SRADemoChainLaink}
\end{figure}

%
%
\paragraph{\bf How the raw data is reinforced using the SRA block.} 
To illustrate how the raw data is reinforced in practice, we use 2D toy datasets with the objective of facilitating the visualization.
We consider first the following function:
\begin{equation}
\label{ToyF1}
    F_1(x)  = 5x_1-5x_2\mathbbm{1}_{x_1>0} \quad \text{and}\quad  y = \{\mathbbm{1}_{p>0.5} \quad \text{with} \quad p = 1/(1+e^{-F_1(x)}  \}
\end{equation}
As simple as it may seem, this function cannot be directly modeled with a linear model due to the term $\mathbbm{1}_{x_1>0}$, which forces $x_2$ to have no effect on the output when $x_1<0$.
Using the reinforced version of the raw inputs helps to alleviate this problem; as shown in Fig~\ref{SRADemo}.
Fig \ref{fig:SRADemooriginal} shows the original data distribution, with the yellow color indicating the class of interest. In Fig~\ref{fig:SRADemoSra}, we show the representation learned by multiplying the raw inputs with the SRA coefficients. The green color represents a possible decision boundary to separate the two classes.
Through multiplication, values of $x_2$ are significantly reduced (e.g., to 0) when needed (i.e., $o_2 \sim 0$ when $x_1<0$, $x_2<0$), which makes the classes easy to separate with the downstream linear model.\\
\indent We  included another synthetic dataset, the 2D chainLink \cite{Clustering_2D_datasets}, as depicted in Fig\ref{SRADemoChainLaink}. By applying SRA coefficients to this dataset, we acquired a new data representation that enables easy separation of the classes, as shown in Fig\ref{SRADemoChainLinkSRA}. Even without knowledge of the true data generating process, it is apparent that all the purple observations have been moved strategically so that a simple rule, $o_2 > 0$, can effectively isolate nearly all the yellow observations of interest. For a more detailed depiction of the reinforced vectors, please refer to the supplementary materials provided in Section \ref{additionReinforcementProcessViz}.

\paragraph{\bf Can the SRALinear find important features.}
In order to interpret machine learning models, it is essential to perform feature attribution, which involves identifying which variables contributed to a high output score. For classical state-of-the-art models like XGBoost, post-hoc explanation tools such as TreeSHAP and LIME are often used to provide individual prediction explanations. However, these tools can introduce their own biases \cite{ProblemSHAP,amoukou2022accurate}.
In this investigation, we aim to assess SRALinear's ability to identify crucial features in comparison to that of Logistic Regression, a self-explanatory model, and XGBoost coupled with TreeSHAP \cite{TreeSHAP}.  As TreeSHAP calculates the exact Shapley value and is computationally efficient, it is particularly well-suited for tree-based models.
For this purpose, we  generate two synthetic datasets with 5 features $ \mathbf x = (x_1, x_2, x_3, x_4, x_5)$ of size 30000 and 60000, respectively based on the Gaussian distribution (0 mean and variance 1) as follows:
\begin{equation}\label{AccurateShap}
    y  = (5x_1-5x_2)\mathbbm{1}_{x_5\le 0}+(5x_3-5x_4)\mathbbm{1}_{x_5> 0}
    \vspace{-0.5cm}
\end{equation}
  
\begin{equation}\label{imbalancedRules}
y =  1 \quad  \text{if} \quad (x_1+2.5)^2+x_2^2<1 \quad \text{or} \quad (x_1-2.5)^2+(x_2-1.5)^2<1 \quad \text{and}  \quad 0   \quad \text{otherwise}
\end{equation}
\indent The example called \textit{Synthetic 1} (Equation \ref{AccurateShap}) is borrowed from \cite{amoukou2022accurate} . It is interesting for this work because it highlights the interactions between the features. The goal is to design a model that can achieve perfect accuracy by using only the features $x_1$ and $x_2$, or alternatively, depending on the sign of $x_5$, using only the features $x_3$ and $x_4$. To evaluate the model's performance, we compute the True Positive Rate (TPR) using a test set consisting of 20\% of the data points, with the remaining 80\% used for training. We restrict our analysis to those data points with $x_5\le 0$, which comprise 3750 instances. Specifically, we assess the ability of SRALinear to identify the two most important features among $(x_1, x_2, x_3, x_4)$.
Regarding the example that we called  \textit{Synthetic 2} (Equation \ref{imbalancedRules}), it is rather attribution tools friendly as features are independent (although there is a non-linearity). Only $x_1$ and $x_2$ are relevant to predict the class 1. We consider all data points from class 1 in test set (695 data points) and  try to find the two most important  features among $(x_1, x_2, x_3, x_4, x_5)$.\\
\begin{table}[t]
\begin{center}
\caption{Relevance feature discovry capacity. The True Positive Rate (TPR) (\%) is  used as metric. $R^2$ (the higher the better) is to evaluate the test performance of \textit{Synthetic 1} and AUCROC (\%) is used for  \textit{Synthetic 2} dataset.}
\label{tab:ComputeImportance}
    \begin{tabular}{|c|c|c|c|}
    \hline
   Datasets & Models & TPR  & Test performance  \\
    \hline
     \multirow{3}{6em}{\textit{Synthetic 1}} & Linear Regression & 51.28 & 50.00\\
     & XGBoost+TreeSHAP & 75.47 &  99.21\\
    & SRALinear & 99.77 &  99.67\\
     \hline
     \hline 
     \multirow{3}{6em}{\textit{Synthetic 2}}& Logistic Regression (LR) & 66.83 & 73.77\\
     & XGBoost+TreeSHAP & 98.63 &  99.99\\
     & SRALinear & 99.86 &  99.72\\
     \hline
    \end{tabular}
\end{center}
\vspace{-8mm}
\end{table}
\indent As shown in Table \ref{tab:ComputeImportance}, SRALinear is able to accurately detect the most relevant features with a high True Positive Rate (TPR) of over 99\%. As expected, TreeSHAP (combined with XGBoost) is able to accurately detect the two most relevant features for \textit{Synthetic 2}, but struggles with the Synthetic 1 dataset, achieving a TPR of approximately 75\%.
Knowing that XGBoost has a perfect performance on this dataset ($R^2 >99\%$), we argue that the incorrect attributions are due to the interpretability tool, which fails to provide the important features in XGBoost's decision. For brevity, we encourage interested readers to refer to \cite{amoukou2022accurate,ProblemSHAP,huang2023inadequacy} for more details on attribution methods and variable interactions.
Regarding the Linear models (Linear regression and Logistic Regression), although being highly explainable, they detect important variables only with moderate accuracy. This is due to bias  when using linear models for handing non-linear data ($R^2 = 50$ \%, AUCROC = 74\%).
From these two synthetic examples, we show that there are two possible biases when using feature attribution: (i) the first is due to underfitting (e.g., using linear models to fit complex data);
(ii) the second is due to post-hoc interpretabilty tools used to explain full complexity models.
In the context mentioned above, the SRALinear model appears to be a good compromise for both the feature attribution and accuracy aspects.

\begin{figure}[t]
    \begin{subfigure}{0.50\textwidth}
        \includegraphics[width = 0.90\textwidth,height=3cm]{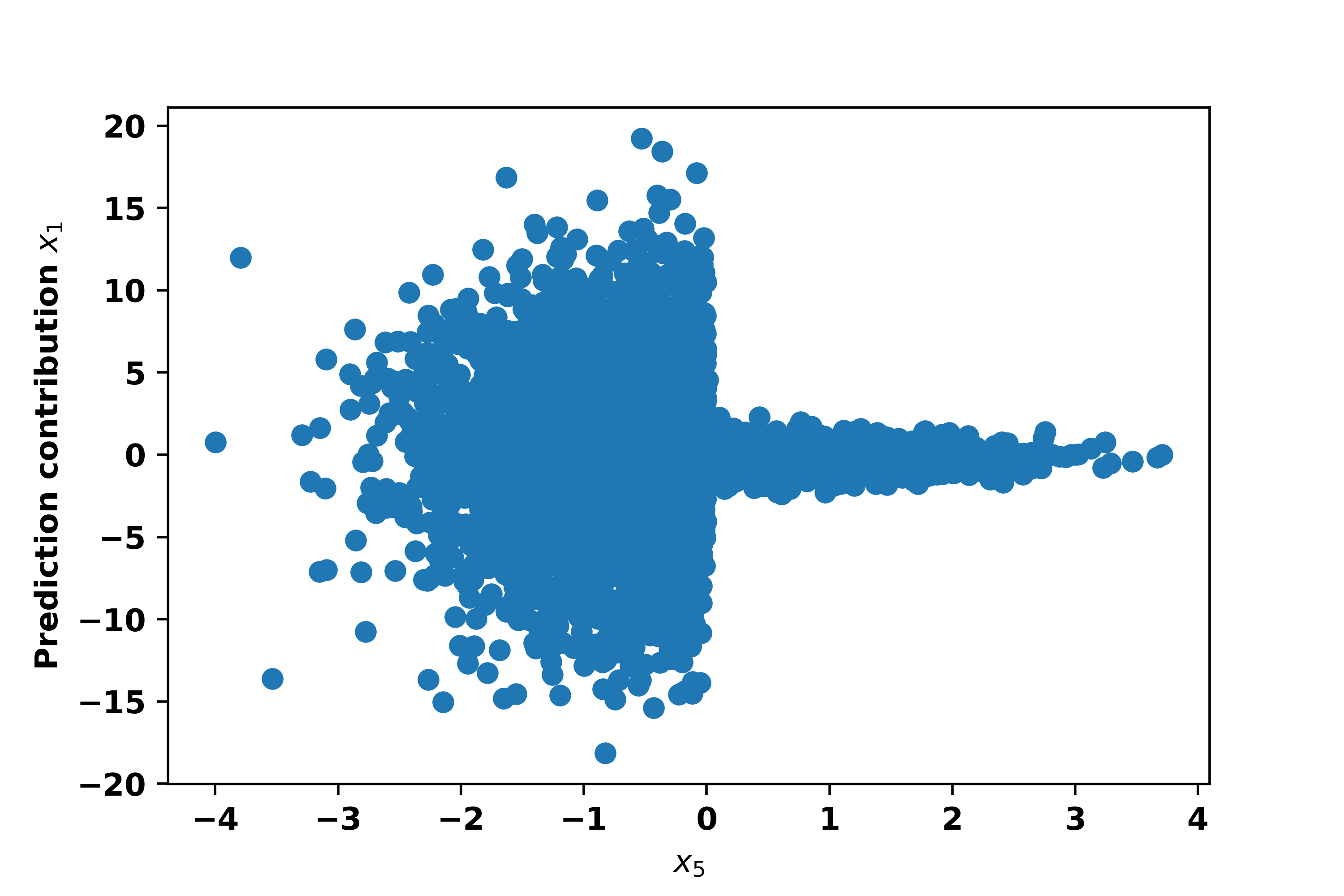}
        \caption{Prediction importance of $x_1$ vs $x_5$}
    \end{subfigure}
    \begin{subfigure}{0.5\textwidth}
        \includegraphics[width = 0.90\textwidth,height=3cm]{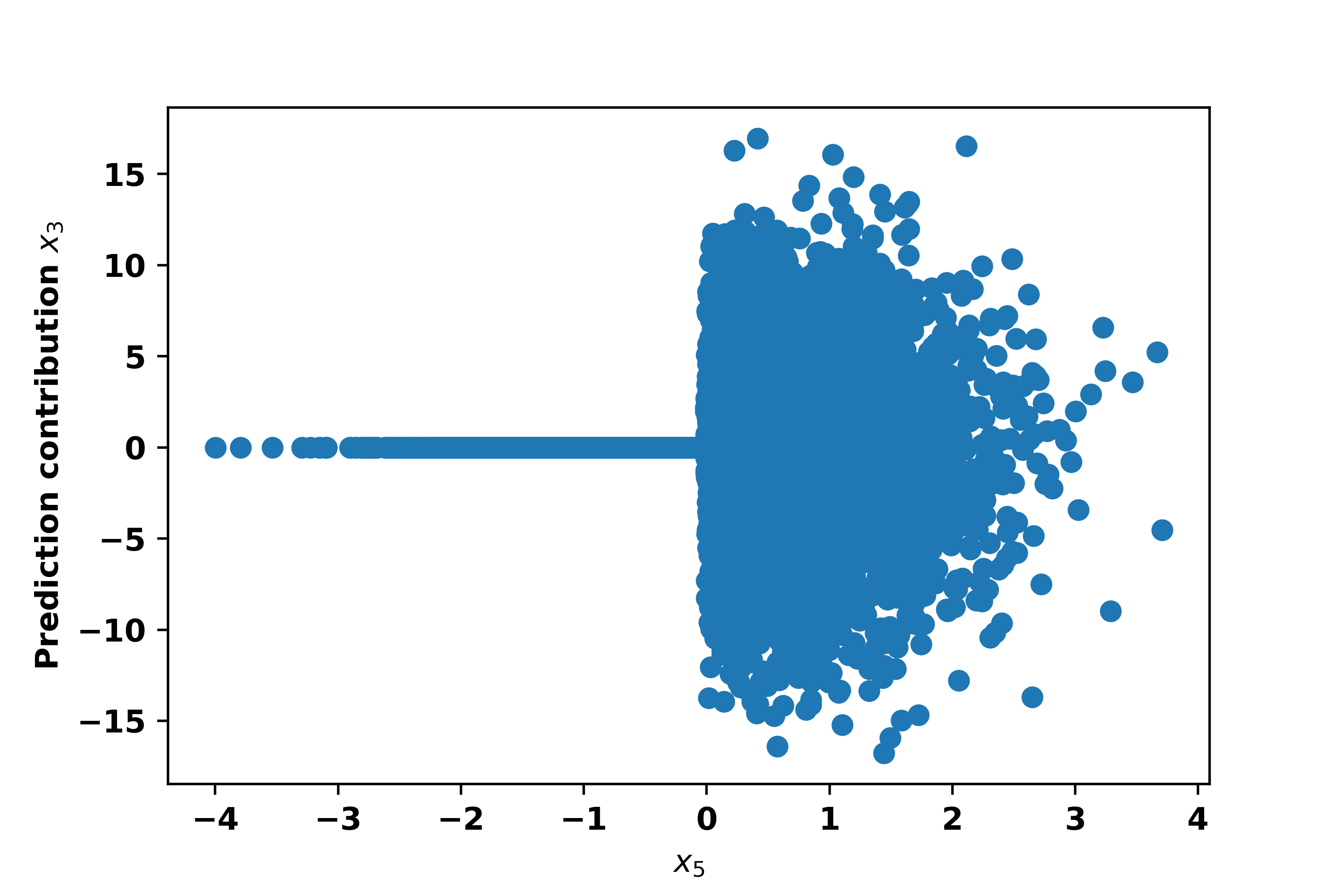}
        \caption{Prediction importance of $x_3$ vs $x_5$}
    \end{subfigure}
    \caption{\textit{synthetic 1}: relevance analysis of the feature $x_5$}
    \label{SRALimite}
\end{figure}

\paragraph{\bf Limitations of SRA based explanations.}
The SRA model, as proposed, should not be used directly as a global feature selector but rather after identifying all relevant variables. This is because the feature importance measure provided by SRALinear is 'the local prediction importance' and not 'the local feature importance' (cf. Equation \ref{eqn:SRA}). Although  these two terms are usually used interchangeably in the literature of feature attribution methods,  there are some nuances \cite{lemhadri2022rbx}. Specifically, the feature that is important to a prediction is automatically relevant, but the inverse is not always true, especially when there are interactions. Regarding the SRALinear model, an illustrative example is the \textit{synthetic 1} dataset (Equation \ref{AccurateShap}). For this dataset, a perfect SRALinear model will always give zero prediction importance to the feature $x_5$ as it cannot be used as main effect or feature (although it can be used to reduce the contribution of other features in the attention vector). Thus, based solely on the prediction importance, one may be tempted to delete the feature $x_5$ to create a model with fewer variables. However with further analysis (e.g., visualizing or computing the gradient $\beta_ia_ix_i \text{ vs } x_5$) we can notice that $x_5$ must be kept.
An shown in Fig~\ref{SRALimite}, an important information needs to be known about $x_5$; which is its sign. When $x_5<0$ (resp. $x_5>0$), the prediction contribution (or importance) of $x_3$ is close to 0 (resp. the prediction contribution of $x_1$ is close to 0). 
A similar visualization would lead to the same finding for the contributions of $x_2$ and $x_4$, indicating that $x_5$ is indeed relevant to the model. Dropping it would result in a drastic reduction in SRALinear's performance, as it would behave like a simple linear regression.
%
\subsection{The effectiveness of the SRA block.}
\label{SRAeffectiveness}
In this section, we discuss the effectiveness of considering the SRA block by comparing the accuracy achieved by SRALinear model (Equation \ref{eqn:SRA}) on benchmark datasets relatively to baseline models (interpretable and non-counterparts).\\
\paragraph{\bf Baseline models.}
We compare quantitatively the performance of the proposed SRA models (Equation \ \ref{eqn:SRA}) with the following baselines:
\begin{itemize}
    \item Logistic Regression (LR): It is a highly interpretable model obtained by simple linear combination of features followed by Sigmoid activation for binary classification problems.
    \item  MultiLayer Perceptron (MLP): it is a full complexity architecture that can model non-linear effects and interactions; making it not directly interpretable by humans.  
    We consider 
    two (2) hidden layers MLP model of dimensions $\{4\times p, 2\times p\}$ as in \cite{TabTransformer}. $p$ is the input feature dimension.
%
    \item TabNet \cite{Tabnet}: it is a deep learning model that provides  local explanation of its predictions without imposing a limit on the order of interactions between the variables in contrast to \cite{NAM,NODE_GAM}. 
    \item XGBoost \cite{Xgboost}:   
Despite the need for feature attribution tools such as TreeSHAP \cite{TreeSHAP} and LIME \cite{LIME} to explain its local predictions, XGBoost remains a favorite and leading state-of-the-art model for several real-life use cases and tabular learning competitions. 
It is selected for comparison with the intention of measuring the performance that may be lost by preferring an intrinsically interpretable model. It is also to be noted that we do not compare directy to some attention-based models, such as \cite{SAINT,NPT,TabTransformer,FTransformer} as they are more motivated by performance than interpretability and XGBoost can give an idea of the upper bound that these models can reach in most cases. 
\end{itemize}
%
%
%
\begin{table}[t]
\begin{center}
\caption{Accuracy of the SRALinear model. Mean and standard deviation AUC (\%), reported from 5-stratified cross validation. Bold highlights the best performance when comparing self-explainable (LR, TabNet, SRALinear) models and italic is used for the overall best performing model.}
\label{tab:compareall}
    \begin{tabular}{|c|c|c|c||c|c|}
    \hline
  Datasets & LR & TabNet & SRALinear & MLP & XGBoost\\
    \hline
    BankChurn & 76.93 (1.56) &  \textbf{86.99} (0.79) &  {86.98} (0.46) & \textit{87.08} 
 (0.73) &  {86.82} (0.79) \\
     \hline 
    CreditDefault & 72.53 (0.49) &  \textbf{77.85} (1.03) & { 77.55} (0.56) & {78.24} (0.78)  & \textit{78.56} (0.69)\\
     \hline 
     BankMarketing   &  90.79 (0.49) & 92.74 (0.70) & {\bf 93.33} (0.50) &  93.44 (0.41)  &   \textit{93.82 } (0.38)\\
     \hline 
    AdultIncome  &  90.50 (0.41) & 90.46 (0.52) & {\bf 91.07} (0.42)  & 91.45 (0.38)   & \textit{92.63} (0.37) \\
     \hline   
    CreditCardFraud &  77.08 (2.59)  & 81.09 (3.92)  &  \textit{\bf 86.58} (2.81)  & { 85.69} (2.53)   &  {86.54} (2.19) \\
  \hline
  Blastchar & 84.54 (1.48) & 83.53 (1.45) & \textbf{84.63} (1.51) & {84.63 } (1.52) & \textit{84.89} (1.21) \\
  \hline 
  TelcoChurn & 88.95 (0.29) & 90.45(0.33) &  \textbf{90.52} (0.31)  &   {90.54} (0.28) &  \textit{91.13} (0.37) \\
  \hline 
  HelocFico & 78.26 (0.52) & 79.39 (0.57) & \textbf{79.43} (0.41) & {79.50} (0.46)  & \textit{79.75} (0.74) \\
    \hline
    \end{tabular}
\end{center}
\end{table}

\paragraph{\bf Evaluation of the Accuracy.}
As shown in Table \ref{tab:compareall}, SRALinear achieved the best performance in 6/8 cases among self-explainable models (over TabNet, LR). Furthermore, the obtained performance is often close (for 6/8 benckmark datasets) to the one of the overall best performing model which is XGBoost. 
 These results confirms the effectiveness of  SRA block  particularly when observing the difference of performances between the Logistic Regression (LR) and SRALinear which ranges from +0.09 for the AdultIncome dataset to +10.05 AUC for the Bank Churn dataset.  We recall that LR model is the resulting architecture when removing the SRA block or setting attention weights to 1 (cf. Fig~\ref{SRAarchi}). 

\vspace{-0.3cm}
\section{Conclusion}
\label{ConclusionSection}
We  presented a novel attention mechanism for tabular learning named 
Self-Reinforcement Attention (SRA), 
a deep learning based representation learning block to produce reinforced version from raw input data through element-wise multiplication. We  demonstrated the effectiveness and the benefits of  SRA  with both synthetic and benchmark imbalanced classification datasets.
We also showed that the SRA models are intelligible in sense that they provides an intrinsic  attribution for feature, which can be further used for global model behavior understanding.
Our experimental results confirms the proposed model as a promising solution for self-explainable models in tabular learning settings without the need to 'sacrificing the accuracy'. 
Overall, we recommend to the interested user to check as much as possible the agreement of the SRA based explanations with their data knowledge since these are not causalities.
The SRA block as proposed can be further enriched especially to deal with complex
tasks. In this concern, we are currently working on how to use several heads and layers,
similar to what is often done in attention-based architectures.
Also, studying empirically the local stability of SRA explanations is an important direction of future research as well as incorporating data knowledge in the training phase (e.g. use monotonic constraints with respect to some features). 

%
%
%
\bibliographystyle{splncs04}
\bibliography{mybibliography}

\begin{thebibliography}{10}
\providecommand{\url}[1]{\texttt{#1}}
\providecommand{\urlprefix}{URL }
\providecommand{\doi}[1]{https://doi.org/#1}

\bibitem{NAM}
Agarwal, R., Melnick, L., Frosst, N., Zhang, X., Lengerich, B., Caruana, R.,
  Hinton, G.E.: Neural additive models: Interpretable machine learning with
  neural nets. Advances in Neural Information Processing Systems  \textbf{34},
  4699--4711 (2021)

\bibitem{amoukou2022accurate}
Amoukou, S.I., Sala{\"u}n, T., Brunel, N.: Accurate shapley values for
  explaining tree-based models. In: International Conference on Artificial
  Intelligence and Statistics. pp. 2448--2465. PMLR (2022)

\bibitem{Tabnet}
Arik, S.{\"O}., Pfister, T.: Tabnet: Attentive interpretable tabular learning.
  In: Proceedings of the AAAI Conference on Artificial Intelligence. vol.~35,
  pp. 6679--6687 (2021)

\bibitem{RandomForest}
Breiman, L.: Random forests. Machine learning  \textbf{45}(1),  5--32 (2001)

\bibitem{NODE_GAM}
Chang, C.H., Caruana, R., Goldenberg, A.: Node-gam: Neural generalized additive
  model for interpretable deep learning. arXiv preprint arXiv:2106.01613
  (2021)

\bibitem{Xgboost}
Chen, T., Guestrin, C.: Xgboost: A scalable tree boosting system. In:
  Proceedings of the 22nd acm sigkdd international conference on knowledge
  discovery and data mining. pp. 785--794 (2016)

\bibitem{aucrocVsaucpr}
Davis, J., Goadrich, M.: The relationship between precision-recall and roc
  curves. In: Proceedings of the 23rd international conference on Machine
  learning. pp. 233--240 (2006)

\bibitem{dilmi2023epigenetics}
Dilmi, M.D., Azzag, H., Lebbah, M.: Epigenetics algorithms:
  Self-reinforcement-attention mechanism to regulate chromosomes expression
  (2023)

\bibitem{FTransformer}
Gorishniy, Y., Rubachev, I., Khrulkov, V., Babenko, A.: Revisiting deep
  learning models for tabular data. Advances in Neural Information Processing
  Systems  \textbf{34},  18932--18943 (2021)

\bibitem{TabTransformer}
Huang, X., Khetan, A., Cvitkovic, M., Karnin, Z.: Tabtransformer: Tabular data
  modeling using contextual embeddings. arXiv preprint arXiv:2012.06678  (2020)

\bibitem{huang2023inadequacy}
Huang, X., Marques-Silva, J.: The inadequacy of shapley values for
  explainability. arXiv preprint arXiv:2302.08160  (2023)

\bibitem{AttentioNotExplantion}
Jain, S., Wallace, B.C.: Attention is not explanation. arXiv preprint
  arXiv:1902.10186  (2019)

\bibitem{LightGBM}
Ke, G., Meng, Q., Finley, T., Wang, T., Chen, W., Ma, W., Ye, Q., Liu, T.Y.:
  Lightgbm: A highly efficient gradient boosting decision tree. Advances in
  neural information processing systems  \textbf{30} (2017)

\bibitem{NPT}
Kossen, J., Band, N., Lyle, C., Gomez, A.N., Rainforth, T., Gal, Y.:
  Self-attention between datapoints: Going beyond individual input-output pairs
  in deep learning. Advances in Neural Information Processing Systems
  \textbf{34},  28742--28756 (2021)

\bibitem{ProblemSHAP}
Kumar, I.E., Venkatasubramanian, S., Scheidegger, C., Friedler, S.: Problems
  with shapley-value-based explanations as feature importance measures. In:
  International Conference on Machine Learning. pp. 5491--5500. PMLR (2020)

\bibitem{lemhadri2022rbx}
Lemhadri, I., Li, H.H., Hastie, T.: Rbx: Region-based explanations of
  prediction models. arXiv preprint arXiv:2210.08721  (2022)

\bibitem{TreeSHAP}
Lundberg, S.M., Erion, G., Chen, H., DeGrave, A., Prutkin, J.M., Nair, B.,
  Katz, R., Himmelfarb, J., Bansal, N., Lee, S.I.: From local explanations to
  global understanding with explainable ai for trees. Nature machine
  intelligence  \textbf{2}(1),  56--67 (2020)

\bibitem{NODE}
Popov, S., Morozov, S., Babenko, A.: Neural oblivious decision ensembles for
  deep learning on tabular data. arXiv preprint arXiv:1909.06312  (2019)

\bibitem{LIME}
Ribeiro, M.T., Singh, S., Guestrin, C.: " why should i trust you?" explaining
  the predictions of any classifier. In: Proceedings of the 22nd ACM SIGKDD
  international conference on knowledge discovery and data mining. pp.
  1135--1144 (2016)

\bibitem{SAINT}
Somepalli, G., Goldblum, M., Schwarzschild, A., Bruss, C.B., Goldstein, T.:
  Saint: Improved neural networks for tabular data via row attention and
  contrastive pre-training. arXiv preprint arXiv:2106.01342  (2021)

\bibitem{Dropout}
Srivastava, N., Hinton, G., Krizhevsky, A., Sutskever, I., Salakhutdinov, R.:
  Dropout: a simple way to prevent neural networks from overfitting. The
  journal of machine learning research  \textbf{15}(1),  1929--1958 (2014)

\bibitem{Clustering_2D_datasets}
Ultsch, A.: Clustering wih som: U* c. Proc. Workshop on Self-Organizing Maps
  (01 2005)

\bibitem{VanillaTransformer}
Vaswani, A., Shazeer, N., Parmar, N., Uszkoreit, J., Jones, L., Gomez, A.N.,
  Kaiser, {\L}., Polosukhin, I.: Attention is all you need. Advances in neural
  information processing systems  \textbf{30} (2017)

\end{thebibliography}
\newpage
\appendix
\section{Additional experimentals results}
\subsection{How the raw data is reinforced using the SRA block.}
\label{additionReinforcementProcessViz}
\begin{figure}[H]
    \begin{subfigure}{0.50\textwidth}
        \includegraphics[width = 0.99\textwidth]{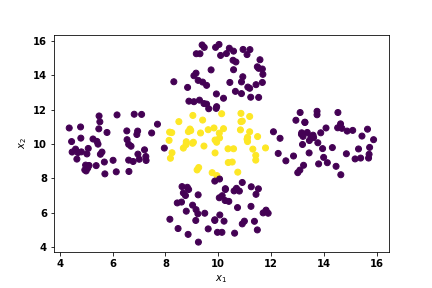}
        \caption{{Original input data}}
                \label{SRADemoFiveSphereOriginal}
    \end{subfigure}
    \begin{subfigure}{0.5\textwidth}
        \includegraphics[width = 0.99\textwidth]{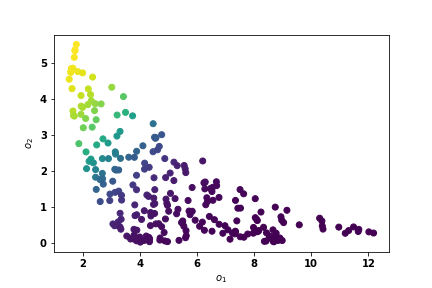}
        \caption{{Learned reinforced inputs with SRA}}
        \label{SRADemoFiveSphereSRA}
    \end{subfigure}
    \caption{Illustration on  Five sphere with 250 datapoints}
    \label{SRADemoFiveSphere}
    \vspace{-1cm}
\end{figure}
\begin{figure}[H]
    \begin{subfigure}{0.50\textwidth}
        \includegraphics[width = 0.99\textwidth]{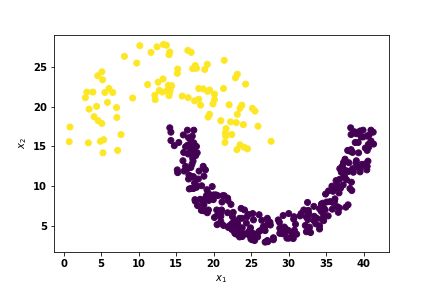}
        \caption{{Original input data}}
        \label{SRADemotwommonOrginal}
    \end{subfigure}
    \begin{subfigure}{0.5\textwidth}
        \includegraphics[width = 0.99\textwidth]{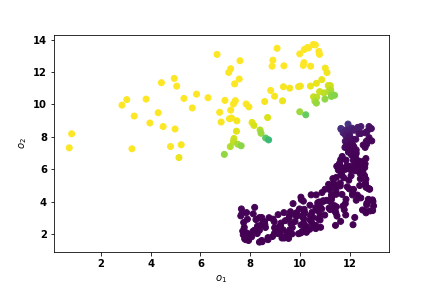}
        \caption{{Learned reinforced inputs with SRA}}
        \label{SRADemotwommonSra}
    \end{subfigure}
    \caption{Illustration on  two moon with 373 datapoints}
    \label{SRADemotwommon}
    \vspace{-0.8cm}
\end{figure}
\begin{figure}[H]
    \begin{subfigure}{0.50\textwidth}
        \includegraphics[width = 0.99\textwidth]{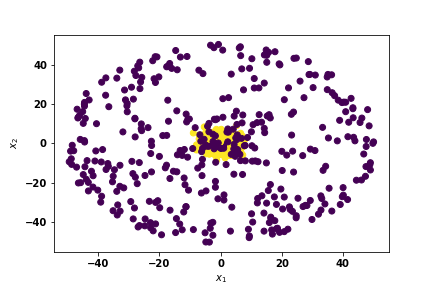}
        \caption{{Original input data}}
                \label{SRADemoTwodiskOriginal}
    \end{subfigure}
    \begin{subfigure}{0.5\textwidth}
        \includegraphics[width = 0.99\textwidth]{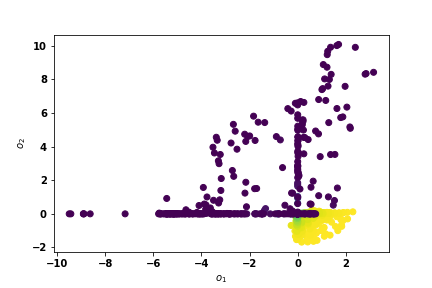}
        \caption{{Learned reinforced inputs with SRA}}
        \label{SRADemoTwodiskSRA}
    \end{subfigure}
    \caption{Illustration on two disks with 800  datapoints}
    \label{SRADemoAtom}
    \vspace{-0.8cm}
\end{figure}
\begin{figure}[H]
    \begin{subfigure}{0.50\textwidth}
        \includegraphics[width = 0.99\textwidth]{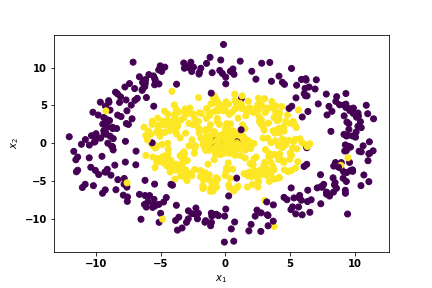}
        \caption{{Original input data}}
                \label{SRADemoringsOriginal}
    \end{subfigure}
    \begin{subfigure}{0.5\textwidth}
        \includegraphics[width = 0.99\textwidth]{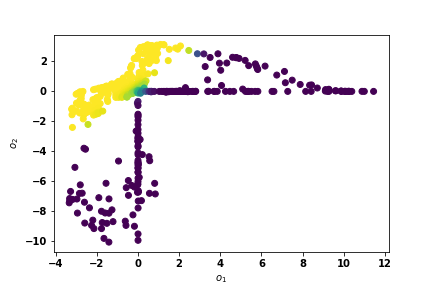}
        \caption{{Learned reinforced inputs with SRA}}
        \label{SRADemoringsSRA}
    \end{subfigure}
    \caption{Illustration on  Rings with 1000  datapoints}
    \label{SRADemorings}
    \vspace{-0.8cm}
\end{figure}
\begin{figure}[H]
    \begin{subfigure}{0.50\textwidth}
        \includegraphics[width = 0.99\textwidth]{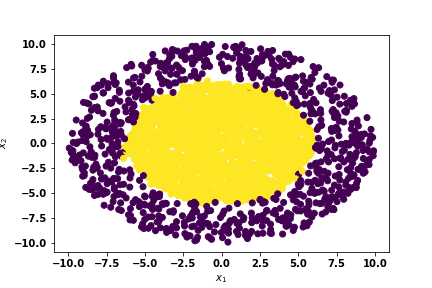}
        \caption{{Original input data}}
                \label{SRADemoDenseDiskOriginal}
    \end{subfigure}
    \begin{subfigure}{0.5\textwidth}
        \includegraphics[width = 0.99\textwidth]{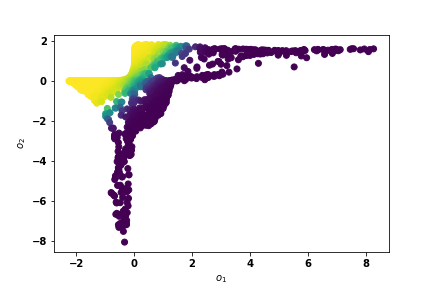}
        \caption{{Learned reinforced inputs with SRA}}
        \label{SRADemoDenseDiskSRA}
    \end{subfigure}
    \caption{Illustration on Dense disk 3000  datapoints}
    \label{SRADemoDenseDisk}
\end{figure}
\end{document}